%% file: main.tex
\documentclass[letterpaper, 10 pt, conference]{ieeeconf}  

\IEEEoverridecommandlockouts                              

\usepackage{graphics} 
\usepackage{algorithm}
\usepackage{algorithmic}
 \usepackage{dblfloatfix}
\usepackage{amsfonts,amssymb,amsmath,bm}
\usepackage{enumerate}
\usepackage{cite}
\usepackage[normalem]{ulem}
\usepackage{caption}
\usepackage{adjustbox}
\usepackage{booktabs}

\makeatletter
\newcommand\notsotiny{\@setfontsize\notsotiny\@vipt\@viipt}
\makeatother
\usepackage{xcolor}

\usepackage[font=footnotesize,labelfont=bf,tableposition=top]{caption}
\usepackage{hyperref}
\hypersetup{
    colorlinks,
    linkcolor={red!50!black},
    citecolor={blue!50!black},
    urlcolor={blue!80!black}
}

\title{\LARGE \bf
Active-Perceptive Motion Generation for Mobile Manipulation
}

\author{Snehal Jauhri$^{1}{}^{*}$, Sophie Lueth$^{1}{}^{*}$, and Georgia Chalvatzaki$^{1,2,3}$
\thanks{*Authors contributed equally}
\thanks{This research received funding from the European Union's Horizon program under grant agreement no. 101120823, project MANiBOT, the German Research Foundation (DFG) Emmy Noether Programme (CH 2676/1-1), and the Daimler Benz foundation.}
\thanks{$^{1}$Computer Science Department, Technische Universit\"at Darmstadt, Germany $^{2}$Hessian.AI, Darmstadt, Germany $^{3}$ Center for Mind, Brain and Behavior, Uni. Marburg and JLU Giessen, Germany \{\texttt{snehal.jauhri,georgia.chalvatzaki\}@tu-darmstadt.de}, \texttt{sophie.lueth@stud.tu-darmstadt.de}
}}

\let\oldtwocolumn\twocolumn
\renewcommand\twocolumn[1][]{%
    \oldtwocolumn[{#1}{
        \vspace{-0.9cm}
    \begin{center}
           \includegraphics[width=\textwidth]{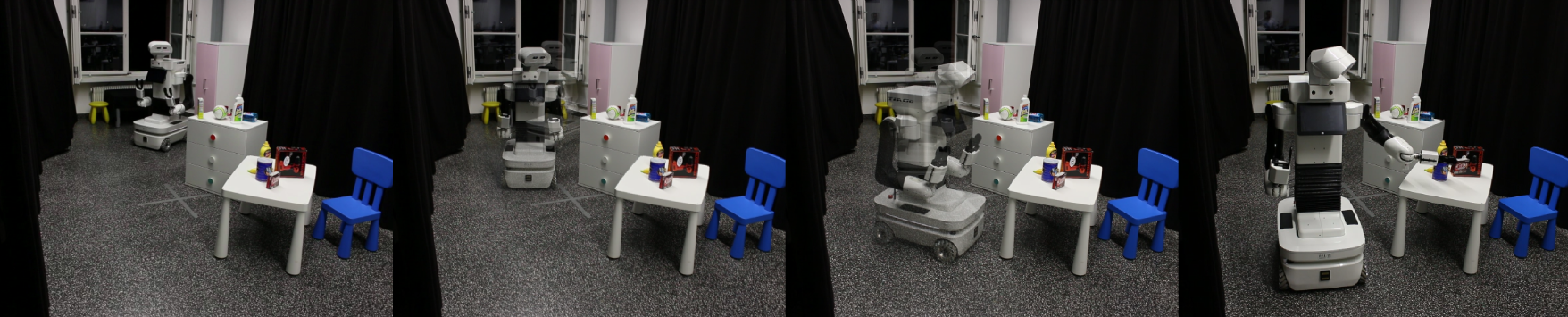}
           \captionof{figure}{Real-world execution of our active perception and grasping pipeline. The mobile manipulator robot, equipped with a head-mounted camera, efficiently explores the scene to detect grasps on the target object and effectively performs the grasp by evaluating its executability. Video demonstrations are provided in the supplementary video and the project website: \href{https://sites.google.com/view/actpermoma/home}{sites.google.com/view/actpermoma}.}
           \label{fig:real_tiago}
        \end{center}
    }]
}

\begin{document}

\maketitle
\thispagestyle{empty}
\pagestyle{empty}

\begin{abstract}
Mobile Manipulation (MoMa) systems incorporate the benefits of mobility and dexterity, due to the enlarged space in which they can move and interact with their environment. However, even when equipped with onboard sensors, e.g., an embodied camera, extracting task-relevant visual information in unstructured and cluttered environments, such as households, remains challenging. In this work, we introduce an active perception pipeline for mobile manipulators to generate motions that are informative toward manipulation tasks, such as grasping in unknown, cluttered scenes. Our proposed approach, Act\textit{Per}MoMa, generates robot paths in a receding horizon fashion by sampling paths and computing path-wise utilities. These utilities trade-off maximizing the visual Information Gain (IG) for scene reconstruction and the task-oriented objective, e.g., grasp success, by maximizing grasp reachability. We show the efficacy of our method in simulated experiments with a dual-arm TIAGo++ MoMa robot performing mobile grasping in cluttered scenes with obstacles. We empirically analyze the contribution of various utilities and parameters, and compare against representative baselines both with and without active perception objectives. Finally, we demonstrate the transfer of our mobile grasping strategy to the real world, indicating a promising direction for active-perceptive MoMa.
\end{abstract}

\input{0_introduction}


\input{2_method}
\input{3_experiments}

\input{4_conclusion}





\clearpage



\bibliographystyle{IEEEtran}
\bibliography{references}
\end{document}

%% file: 0_introduction.tex
\section{Introduction}

\begin{figure*}[t!]
    \centering
    \includegraphics[width=0.95\textwidth]{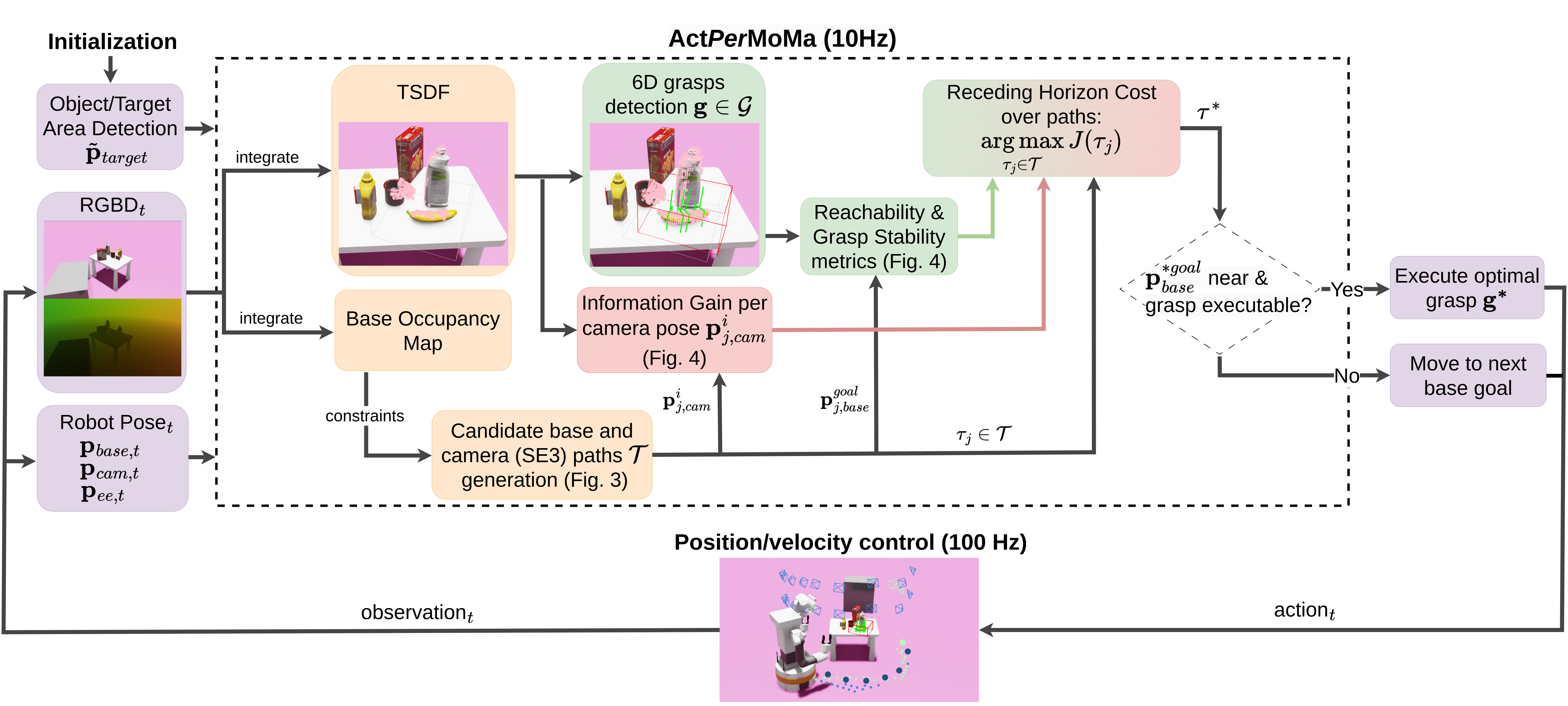}
    \captionof{figure}{Act\textit{Per}MoMa pipeline. Using a rough initial knowledge about the target area or target object position~$\widetilde{\mathbf{p}}_{target}$, we continuously plan and execute informative motions for the mobile grasping task. At every timestep $t$, the RGBD information from the head-mounted embodied camera is integrated into a scene TSDF for both grasp detection and information gain computation. Using the currently known free space for movement of the robot base, we sample candidate robot paths~$\mathcal{T}$, including both base and camera poses, towards the target. For each candidate path $\tau_j \in \mathcal{T}$, we compute the information gained from camera views~$\mathbf{p}_{j,cam}^{i}$ in the path, and the reachability of stable detected grasps from the final base poses~$\mathbf{p}_{j,base}^{goal}$ in the path. We trade-off these objectives with a receding horizon cost~$J_{\tau}$ and take a step of the optimal path~$\tau^*$ for execution at every timestep.
    }
    \label{fig:pipeline}
    \vspace{-0.55cm}
\end{figure*}

We envision a near future where embodied agents, such as mobile manipulators, can operate autonomously in everyday environments like households. However, performing tasks in these environments is challenging due to the unstructured nature and unpredictability of the real world. Hence, some robots will actively gather information about their surroundings through embodied sensors, like an embodied camera whilst operating and changing their environment~\cite{bajcsy2018revisiting}. While current advances in AI and machine learning for robotics have unlocked new capabilities for table-top manipulation~\cite{Breyer22,chi2023diffusion,zeng2021transporter}, or language-driven navigation and manipulation~\cite{huang2023visual,wang2023active,shah2023lm,yokoyama2023adaptive,ahn2022can}, mobile manipulation in unknown (or partially known) scenes poses significant challenges~\cite{mittal2022articulated,xia2021relmogen,pankert2020perceptive}, as the MoMa robot needs to consider both scene reconstruction and task-oriented objectives.

Active perception~\cite{aloimonos1988active,aloimonos2013active} refers to the ability of an agent to ``\textit{know why it wishes to sense, and then chooses what to perceive, and determines how, when and where to achieve that perception},'' ~\cite[p.~178]{bajcsy2018revisiting}. For mobile robots, the robot's objective is typically reconstruction, i.e., obtaining volumetric information about the scene/target object~\cite{vasquez2014volumetric,potthast2014probabilistic,zaenker2021combining,Schmid20}. Many proposed active-perception methods use a Next-Best-View (NBV) strategy~\cite{Isler16,bircher2016receding,Watkins21,Naazare22,bartolomei2021semantic}, primarily choosing viewpoints based on information gain (IG)~\cite{Isler16} that minimizes uncertainty by exploring unobserved regions. A good overview and comparison of different IG formulations for NBV is provided in \cite{Delmerico18}. Notably, an active perception approach that only considers movement to the NBV with the most information gain can lead to unnecessarily large motions. Hence, the authors of~\cite{zaenker2023graph} consider IG over paths using a graph-based approach, while in \cite{Bircher16}, a receding horizon viewpoint and path planning method is proposed since this formulation naturally benefits active perception as it adapts to newly observed information. 

This work focuses on active perception that enables mobile manipulation in unknown environments, with a focus on mobile grasping. When grasping with static manipulators, recent methods adopt grasp quality metrics to choose the next robot viewpoint that minimizes uncertainty in the grasp pose estimation~\cite{Morrison19, Breyer22}. Breyer et al.~\cite{Breyer22} exploit the fact that only negligible performance differences can be detected in the different formulations of volumetric IG~\cite{Isler16, Delmerico18} and propose a rear-side voxel IG computation over views equally distributed on a half-sphere above the target object. The active perception is considered complete when a stable grasp quality has been found. However, these methods have several drawbacks when considering their application to MoMa. First, a MoMa robot can be placed \textit{anywhere} in a scene and can reach \textit{any} viewpoint (subject to scene restrictions). Thus, the range of movements of a MoMa robot is larger which makes wasteful motions especially costly. This also means that NBV-only approaches are sub-optimal as they do not consider information gained \textit{during} movement from one NBV to another. Second, the reachability and feasibility of grasps and viewpoints are crucial when formulating planning for MoMa, since grasps of high quality could be challenging to reach and might cause unnecessary robot movements. 

In this work, we propose an effective and efficient approach for visually informative motion generation for mobile manipulators to perform tasks in unknown, cluttered scenes. Particularly, we consider the problem of \textit{mobile grasping}. Our method, Active-Perceptive Motion Generation for Mobile Manipulation (Act\textit{Per}MoMa), resolves mobile grasping efficiently by planning over paths, collecting enough visual scene information to infer collisions and good grasps, and accounting for grasp executability. We ablate and benchmark our method against active grasping baselines, showcasing design choices that lead to performance gains toward MoMa applications.



To summarize: (i) We propose a novel formulation for active perceptive mobile manipulation, generating paths toward objects of interest without any prior scene knowledge. (ii) We calculate path-wise utilities over robot poses. Our motion generation objective balances exploration, maximizing visual information gain, and exploitation of task-specific information, such as grasp executability for mobile grasping. (iii) To ensure robot reactivity to new data, we use a receding-horizon control approach. We sample and evaluate numerous potential paths towards an approximate object location, based on feasible base goal-poses perceived in the current scene.


%% file: 2_method.tex
\section{Active-Perceptive Motion Generation for Mobile Manipulation}
We consider scenarios where a MoMa robot is placed in a previously unseen environment and is tasked with picking up a target object placed on a surface among clutter. To achieve this task, the mobile manipulator needs to use its multiple embodiments, i.e., a mobile base to move in the scene, an RGBD camera to perceive the scene and gather information, and an arm/end-effector to execute 6DoF grasps. Without loss of generality to different physical MoMa designs, we can simplify the description of the robot's state as a combination of its mobile base pose $\mathbf{p}_{base} \in SE(2)$, its camera pose $\mathbf{p}_{cam} \in SE(3)$, and its end-effector pose $\mathbf{p}_{ee} \in SE(3)$. In principle, we could consider a whole-body joint representation, but in this work we decouple the poses of the different embodiments for simplicity.


Hypothesizing some rough knowledge of the area where the target object is, we obtain an approximate bounding box of the region where the target object lies, the center of which we denoted by $\widetilde{\mathbf{p}}_{target}$. In practice, this can be done by either exploring and executing an RGB object detector or via information from a user instruction such as ``Pickup the object from the right corner of the table'', but no prior information about the scene is assumed. Using the point cloud from the embodied camera we can build a volumetric representation of the scene, a 3D Truncated Signed Distance Function (TSDF), to effectively plan collision-free motions in the observed environment. The TSDF is also used to detect a set of 6DoF grasps $\mathcal{G}=\{\mathbf{g}_i\}_{i=0}^{N_g}$, where $\mathbf{g}\in SE(3)$ of maximum number $N_g$, in the object region using an $SE(3)$ grasp detection network that can process volumetric information, such as \cite{Breyer2020,Jiang2021}. 
Example scenarios we consider are visualized in Fig.~\ref{fig:pipeline} and Fig.~\ref{fig:paths_sim}.

Our overall objective is to generate efficient motions for the MoMa robot to find and execute a grasp to pick up the target object. For mobile grasping with active information accumulation, we consider the following design principles for motion generation:
\\
(i)~Mobile manipulators operating under partial information need to trade-off between exploration (scene understanding) and exploitation (executing a task-oriented action).\\
(ii)~Since information is continuously gathered, the control formulation must be adaptive and reactive, and the costs must be considered and updated at every time interval.\\
(iii)~The key utilities to be balanced are the robot movement cost, the information gained about the target-object/scene, and the likelihood of grasp success.\\
(iv)~While grasp detectors~\cite{Breyer2020,Jiang2021} may predict high-quality grasps, they are not necessarily feasible for the MoMa robot. Thus, grasping utility should consider metrics like the likelihood of reaching a grasp from different base locations~\cite{jauhri2022robot}. 

In this section, we introduce a holistic motion generation pipeline that satisfies these principles (illustrated in Fig.~\ref{fig:pipeline}). We sample many candidate feasible paths for the robot to execute the task while effectively using its embodiments (sec. \ref{sec:candidates}). To ensure that the robot continuously adapts to new information gathered, we choose a sampling-based receding horizon control formulation (sec. \ref{sec:RHC}). Our control formulation balances the objectives of information gain (sec. \ref{sec:IG}) and the utility of grasp executability (sec. \ref{sec:reach}). 

\subsection{Candidate goals \& paths generation}
\label{sec:candidates}
Our objective is to move towards a target object, whose location is only roughly known in an unobserved scene, in the most informative and time/energy efficient manner to grasp it. We measure efficiency w.r.t the total distance traveled, viewpoints visited, and number of failures. At each time step, we sample candidate paths for the robot and evaluate utilities over those paths. To approach the target object, we sample ${N_b}$ base poses near the approximate target object position $\widetilde{\mathbf{p}}_{target}$ within a radius that affords robot reachability~\cite{jauhri2022robot,birr2022oriented}. These base poses serve as goals for our path generation $\{\mathbf{p}_{base}^{goal_i}\}_{i=0}^{N_b}$. We ensure that base goals are collision-free by performing a simple collision-check with the scene's continuously generated TSDF grid or base occupancy map. The resampling of new base goals and paths at every time step ensures feasibility based on new scene information. We also jointly sample SE(3) camera poses at these base goal poses such that the robot always looks at the target area.

\begin{figure}[t!]
    \centering
    \includegraphics[width=0.775\columnwidth]{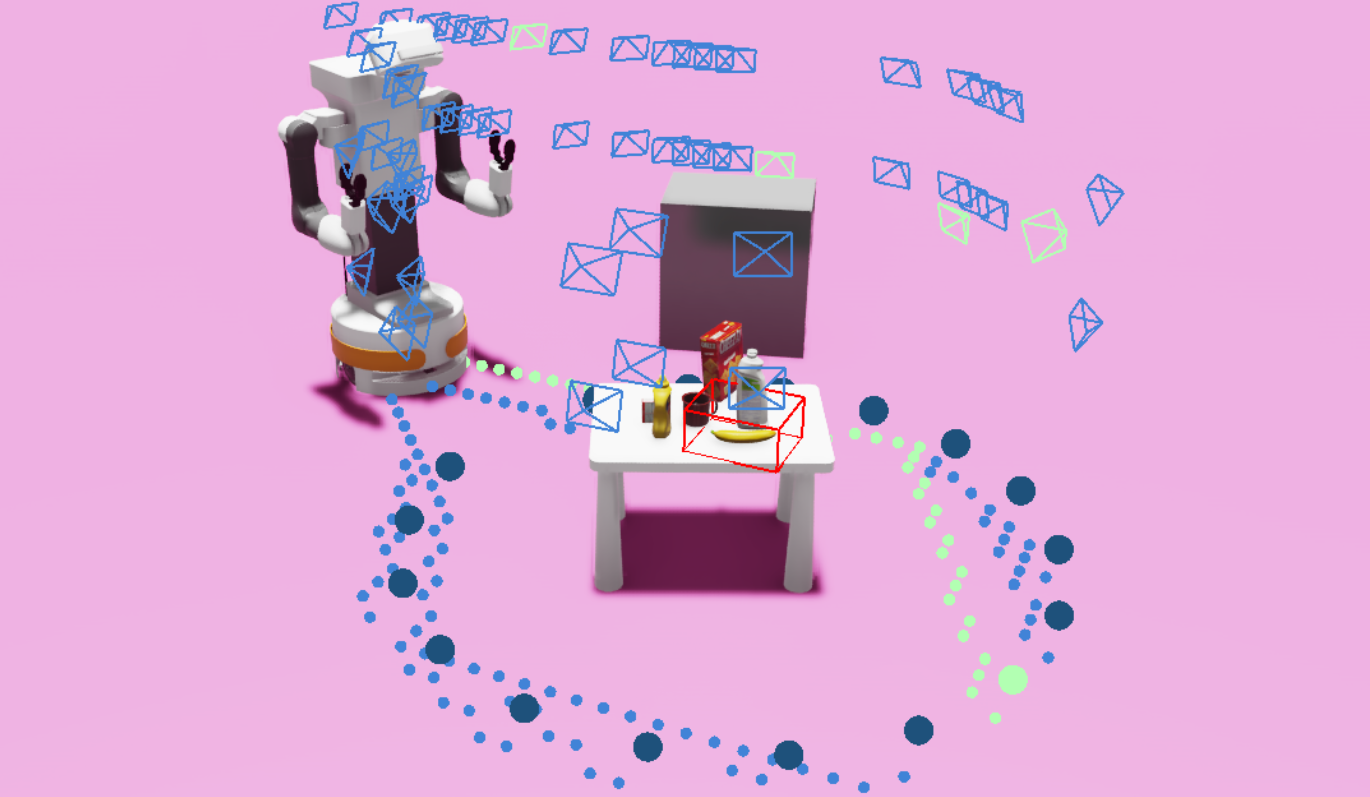}
    \captionof{figure}{Example scene with sampled candidate paths (blue) for the robot pose towards the target object (red box). The paths consist of SE(2) poses for the base and SE(3) poses for the head-mounted camera (visualized from the robot to the base goals). The current optimal path is highlighted in green.}
    \label{fig:paths_sim}
    \vspace{-0.6cm}
\end{figure}

Our objective is to obtain the optimal motion of the robot toward the target object by planning to the candidate base goals, which should allow reaching the object to be grasped. We, thus, sample $M$ candidate paths $\mathcal{T} = \{\tau_j\}_{j=0}^{M}$ to all the $N_b$ candidate base goals $\{\mathbf{p}_{base}^{goal_i}\}_{i=0}^{N_b}$ using optimal path planners---in this work we plan over discretized grids with A$^*$. Each path $\tau \in \mathcal{T}$ consists of base poses from the current robot base to the base goal as well as sampled feasible camera poses $\mathbf{p}_{cam} \in SE(3)$ along the path, i.e.,  $\tau_i = \{ \{\mathbf{p}_{base}^0, \mathbf{p}_{cam}^0\}, \{ \{\mathbf{p}_{base}^1, \mathbf{p}_{cam}^1\} \hdots \{\mathbf{p}_{base}^{goal},\mathbf{p}_{cam}^{goal} \}\}$. Example candidate paths are visualized in Fig. \ref{fig:paths_sim}. We then generate robot motion based on these paths using a receding horizon control formulation, as detailed next.  

\subsection{Receding-horizon control}
\label{sec:RHC}
We use a receding-horizon control formulation to generate the robot's motion to find and execute a grasp on the target object. At each time step, we choose the optimal path among the sampled candidate paths $\mathcal{T}$ and execute an action towards the first waypoint along this current optimal path. The re-computation of actions at every timestep ensures robot reactivity to newly observed scene information.

Formally, given the observation of the scene, i.e., the observed TSDF $\mathbf{o}_{\text{TSDF}}$, the detected set of grasps $\mathcal{G}$, and the sampled candidate paths $\mathcal{T}$, we compute the current optimal path $\tau^* \in \mathcal{T}$ based on the expected information gain ${J_{IG}}$ and the utility of the grasps' executability $J_{exec}$ in the paths:
\begin{align}\label{eq:rhd_optimization}
    \tau^* = \underset{\tau \in \mathcal{T}}{\mathrm{arg\,max}}\, J_{IG}(\mathbf{o}_{\text{TSDF}}, \tau) + J_{exec}(\mathcal{G}, \tau)
\end{align}
Utilities ${J_{IG}}$ and $J_{exec}$ are detailed in subsections \ref{sec:IG} \& \ref{sec:reach}.

For movement at every time step, we use the first waypoint along the chosen optimal path, i.e., $\{\mathbf{p}_{base}^{*1}, \mathbf{p}_{cam}^{*1}\} \in \tau^*$ and run a low-level controller that executes IK-based velocities for the robot base and the camera. If the optimal path $\tau^*$ contains exactly one waypoint, i.e., the robot is close enough to the final chosen base goal $\mathbf{p}_{base}^{*goal}$, we finally consider grasp execution. If the grasp execution utility $J_{exec}$ is above a threshold, we execute the grasp with the highest utility (sec. \ref{sec:reach}) by activating the arm/end-effector and planning a motion to the SE(3) grasp. 




\subsection{Information gain computation \& grasp detection}
\label{sec:IG}

\underline{Information gain computation}: Our perception objective is to obtain more information about the target object in order to grasp it. For this, we use an information gain (IG) formulation inspired by \cite{Delmerico18}, \cite{Breyer22}. We continuously build a voxel-based TSDF representation $\mathbf{o}_{\text{TSDF}}$ of the scene and calculate the rear-side voxel information gain $IG_{rear}$ \cite{Delmerico18}. In this setup, we cast rays from each candidate camera view $\mathbf{p}_{cam} \in SE(3)$ and count voxels that are on the rear side of the observed TSDF voxels and would thus be revealed by the candidate camera view. Since the approximate location of the target object is known, only voxels in an approximate bounding box around the target object are considered. Hence, this IG formulation rewards camera views that see more of the target object than has already been seen. More precisely per~\cite{Breyer22}, for every viewpoint $\mathbf{p}_{cam}$, a set of rays $R$ are generated by casting from a virtual camera placed at the respective view pose. Every ray $r$ traverses voxels of the TSDF $v\subset\mathbf{o}_{\text{TSDF}}$ until it hits an observed surface. Therefore the rear-side IG is computed as $IG_{\text{rear}} = \sum_{r \in R}\sum_v \mathcal{I}(v)$, where $\mathcal{I}(v)=1$ if the voxel is on the rear of an existing voxel and within the approximate target object bounding box.

\begin{figure}[t!]
    \centering    \includegraphics[width=\columnwidth]{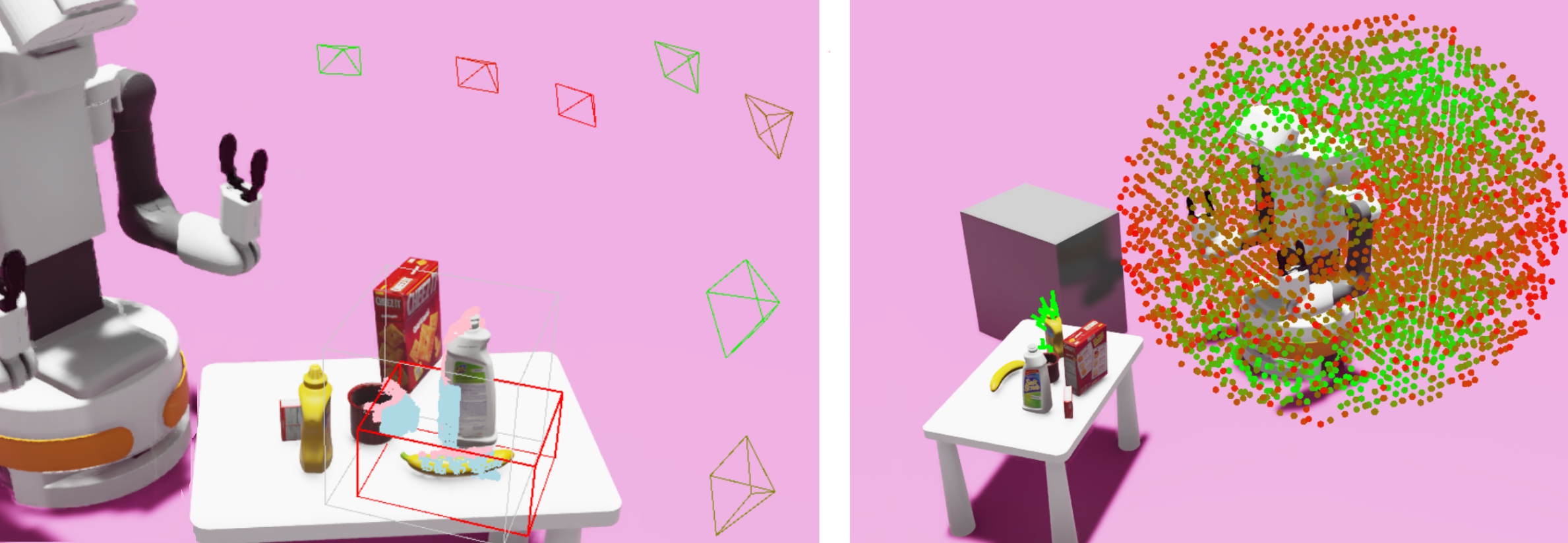}
    \captionof{figure}{\textbf{Left:} Example rear-side Information Gain (IG) for a candidate view. Pink voxels denote observed TSDF voxels. Blue voxels are on the rear side of the observed TSDF, which could be revealed by a candidate view. Views are colored red to green, denoting lower to higher IG. \textbf{Right:} Reachability map of the robot's left arm, reduced from 6 dimensions (SE(3)) to 3 for visualization. Red and green points denote lower and higher reachability. Current detected 6D grasps are visualized in green on a target object.}
    \label{fig:ig_and_reach}
    \vspace{-0.6cm}
\end{figure}

Unlike \cite{Delmerico18}, \cite{Breyer22}, we consider not only the Next-Best-View (NBV) but the IG\textit{ over paths} taken by the robot. Instead of only sampling a few viewpoints around the target object, we consider candidate viewpoints from each candidate path $\tau$ from our sampled paths $\mathcal{T}$. Moreover, we also consider the cost of reaching the viewpoints in the paths by weighting the IG by the distance to the viewpoints~$dist(\mathbf{p}_{cam})$ along the path. This takes care of our requirement that \textit{information gained sooner is better than later}. We can thus calculate the total IG over each candidate path $\tau$ as
\begin{align}\label{eq:ig}
    J_{IG}(\mathbf{o}_{\text{TSDF}}, \tau) = \underset{\mathbf{p}_{cam} \in \tau}{\sum} \frac{IG_{rear}(\mathbf{o}_{\text{TSDF}}, \mathbf{p}_{cam})}{dist(\mathbf{p}_{cam})^2}.
\end{align} 
An example visualization of the rear-side voxel IG is provided in Fig \ref{fig:ig_and_reach}.

\underline{Grasp detection}:
At every time step, we query grasps in the target object region using the observed TSDF of the scene using a grasp detection network. In this work, we use the VGN~\cite{Breyer2020} grasp detection network that can predict an SE(3) grasp pose for every 3D voxel of the TSDF along with a grasp quality prediction~$q$. We use a grasp quality threshold $q_{th}$ hyperparameter to detect good grasps with a high likelihood of success.
Given that the built TSDF contains only partial/incomplete information, it is also important to consider grasp detector inaccuracy. Hence, as in~\cite{Breyer22}, we also consider grasp \textit{stability} by ensuring that a grasp predicted for the same 3D voxel on the TSDF has a high-quality score for $n_{stab}$ steps. Finally, grasps that have a high likelihood of success but are challenging to reach can cause long, sub-optimal robot motions. To this end, we also score grasps based on their reachability from the candidate base goals around the target object, as detailed next.

\subsection{Reachability utility \& grasp selection}
\label{sec:reach}

As we detect stable grasps over time, the motion generation should smoothly switch towards the actual execution of grasps, i.e., switching from exploratory to exploitative behavior. For this, the robot needs to be positioned at a base goal where a high-quality grasp can be executed easily. We achieve this ability by computing a grasp reachability/executability utility $J_{exec}$ corresponding to each candidate path $\tau$.
The reachability of any SE(3) end-effector pose of a robot from a given base pose can be found by pre-computing a reachability map \cite{Zacharias2007, vahrenkamp_robot_2013, makhal_reuleaux_2018}. This map is obtained by executing forward kinematics for many different joint configurations and storing the visitations and the manipulability of the 6D voxels visited by the end-effector. We refer to \cite{Zacharias2007}, \cite{makhal_reuleaux_2018} for a full description of reachability map computation. A visualization of the reachability map used in our approach is provided in Fig \ref{fig:ig_and_reach}.

In our pipeline, we pre-compute the reachability map $\mathcal{R}$ and query it for \textit{each grasp} $\mathbf{g} \in \mathcal{G}$, when executed \textit{from each candidate base goal} $\mathbf{p}_{base}^{goal}$ corresponding to our sampled paths $\tau \in \mathcal{T}$. The highest reachability over all grasps gives us a utility score $J_{exec}$ for each candidate path $\tau$. In the case of a dual-armed mobile manipulator (as used in our experiments) we can also compute the reachability of both arms and use the maximum for utility computation, as well as for arm selection. Moreover, as the robot reaches the optimal base goal, this utility is also used for grasp selection as we execute the most reachable grasp from the final base goal. To ensure that the proximity of the robot to the base goal of the path $\mathbf{p}_{base}^{goal} \in \tau$ is also considered, we weigh the reachability utilities by the length of the path $len(\tau)$, resulting in 
\begin{align}\label{eq:grasp}
    J_{exec}(\mathcal{G}, \tau) =  \frac{\underset{\mathbf{g} \in \mathcal{G}}{\mathrm{max}}\, \mathcal{R}(\mathbf{g}, \mathbf{p}_{base}^{goal})}{len(\tau)}
\end{align}

\subsection{Additional hyperparameters}
\label{sec:algo}
To smoothly switch between the two objectives in (\ref{eq:rhd_optimization}), we weigh the IG and grasp execution utilities by factors $w_{IG}$ and $w_{exec}$. 
To avoid noisy grasps being used for movement and execution, we filter out unstable grasps, i.e., grasps that disappear after a few timesteps. Another problem that can appear is that the robot can oscillate between two base goals that move the robot in opposing directions if they have similar overall utility (\ref{eq:rhd_optimization}). Thus, we introduce a momentum term that continues to move the robot in a direction unless the utility of another direction is significantly higher.

%% file: 3_experiments.tex
\section{Experiments}

\subsection{Experimental setup \& metrics}
We conduct experiments both in simulation and the real world. Our setup contains a dual-armed TIAGo++ MoMa robot with a holonomic base and a head-mounted camera. The robot has 20 DoF in total (2 for the head, 1 for the torso, 3 for the holonomic base, and 7 for each arm), allowing 5D view sampling. In the simulated setup in Isaac Sim, we spawn the robot at a maximum distance of 2~m w.r.t. the approximate target location and consider two scenarios; a simple one, where a table is placed in free space and with 4 randomly spawned objects from the YCB dataset~\cite{calli2015ycb}, and a more complex one with 6 objects to create more clutter and with a random obstacle sampled around the table to obstruct the path of the robot. For grasp detection, we used a pre-trained VGN~\cite{Breyer2020}. The real-world setup resembles the one in simulation as shown in Fig.~\ref{fig:real_tiago}.

We conduct several experiments to ablate our method and justify the selection of key hyperparameters of our algorithm while also comparing against baselines. To measure the performance gains of the compared approaches, we employ the following metrics; \textbf{Success Rate} (SR), the percentage of episodes that finish with successful grasp execution, \textbf{Abort Rate}: (AR) the percentage of episodes that end without finding executable grasps in the given time budget, \textbf{Grasp Failure Rate} (GFR): the percentage of episodes ending in grasp failure, \textbf{total distance covered} ($d_{\text{total}}$), and the \textbf{total number of views} visited ($v_{\text{total}}$)). We run every experiment for each scenario for 500 episodes and report average metrics and standard deviation when applicable. In the tables, we note superior performance of at least 0.5\% in \textbf{bold}.

\begin{table}[t!]
\renewcommand{\arraystretch}{0.9}
\begin{center}
\captionsetup{justification=centering}
  \caption{Ablations \& Hyperparameter study -- Simple scenes}  \label{ablation-simple}
\begin{adjustbox}{width=\linewidth,center}
  \begin{tabular}{@{}l|c|c|c|c|c@{}}
    \toprule
      \multicolumn{6}{c}{Hyperparameters}\\
    \hline \hline
         Approach & SR (\%) $\uparrow$ & AR (\%) $\downarrow$ & GFR (\%) $\downarrow$ & $d_{\text{total}}$ (m) $\downarrow$ & $v_{\text{total}}$ $\downarrow$ \\ \hline
    Act\textit{Per}MoMa-Quality (0.7) & 92.6 & \textbf{1.8} & 5.6 & \textbf{3.93$\pm$1.88} & \textbf{13.81$\pm$6.33}  \\ \hline
    Act\textit{Per}MoMa-Quality (0.9) & \textbf{93.4} & 2.2 & \textbf{4.4} & 4.02$\pm$1.99 & 13.91$\pm$6.31 \\ \hline\hline
    Act\textit{Per}MoMa-StableGrasp (1) & 94.2 & \textbf{1.0} & 4.8 & \textbf{4.01$\pm$1.99} &   \textbf{13.87 $\pm$ 6.22} \\ \hline
    Act\textit{Per}MoMa-StableGrasp (5) & 94.0 & 3.2 & \textbf{2.8} & 4.19$\pm$1.87 & 14.38$\pm$5.87  \\ \hline\hline
    Act\textit{Per}MoMa-IGweight (3.0) & 91.0 & 5.6 & 3.4 & 4.18$\pm$2.21 & 14.36 $\pm$ 6.70 \\ \hline
    Act\textit{Per}MoMa-IGweight (0.2) & \textbf{95.4} & \textbf{1.4} & \textbf{3.2} & \textbf{3.59$\pm$1.69} & \textbf{12.67 $\pm$ 5.39}  \\ \hline \hline
    Act\textit{Per}MoMa-momentum (0) & \textbf{94.8} & 3.2 & \textbf{2.0} & 3.85$\pm$1.92 & 13.38 $\pm$ 5.78  \\ \hline 
    Act\textit{Per}MoMa-momentum (700) & 93.0 & \textbf{2.2} & 4.8 & 3.82$\pm$1.88 & 13.32 $\pm$ 6.00  \\ \hline  
    \midrule
   \multicolumn{6}{c}{Ablation} \\ \hline \hline
    Act\textit{Per}MoMa$^\dagger$  & 95.4 & 1.4 & 3.2 & 3.59$\pm$1.70 & 12.67$\pm$5.39  \\ \hline
    Act\textit{Per}MoMa-IG-only & \textbf{98.4} & \textbf{0.8} & \textbf{0.8}  & \textbf{2.81$\pm$1.06} & \textbf{10.42$\pm$4.0}  \\ \hline
    Act\textit{Per}MoMa-no-weights & 42.4 & 54.6 & 3.0 & 7.96$\pm$2.15 & 25.36$\pm$6.38  \\ \hline
\midrule
    \multicolumn{6}{c}{Hard Grasps}\\ \hline \hline
    Act\textit{Per}MoMa$^\dagger$  & \textbf{66.8}  & \textbf{23.6} & 9.6 & 5.33$\pm$2.77 & 18.75$\pm$9.04  \\ \hline
    Act\textit{Per}MoMa-IG-only & 65.2 & 31.6 & \textbf{3.2} & \textbf{3.59$\pm$2.35} & \textbf{17.88$\pm$9.52} \\ 
    \bottomrule
  \end{tabular}
  \end{adjustbox}
  \end{center}
  \vspace{-0.15cm}
 \scriptsize{$^\dagger$ Quality=0.8, StableGrasp=1, IGweight=0.2, momentum=800}
 \vspace{-0.45cm}
\end{table}

\subsection{Ablations \& hyperparameter study}\label{sec:ablation}
To justify design decisions and better balance our algorithm's exploration-exploitation dilemma, we conduct an extensive ablation study given the aforementioned metrics. In the following, we present the results for (i)~Act\textit{Per}MoMa-IG-only: ablation without the grasp executability objective in which case we execute a grasp as soon as we are within reach of the object, which in practice is 0.85~m away from the approximate target object; (ii)~Act\textit{Per}MoMa-no-weights: ablation without path-length-related scaling of the utilities (see eqs.~\ref{eq:ig},~\ref{eq:grasp})). Additionally, we tune important parameters to find the best configuration for our method and present results for varying grasp quality thresholds~(Act\textit{Per}MoMa-Quality), grasp stability windows~(Act\textit{Per}MoMa-StableGrasp), IG weighting factors~(Act\textit{Per}MoMa-IGweight), and the momentum term that punishes oscillatory paths (Act\textit{Per}MoMa-momentum).

\begin{table}[t!]
\renewcommand{\arraystretch}{0.9}
\begin{center}
    \captionsetup{justification=centering}
  \caption{Ablations \& Hyperparameter study -- Complex scenes}  \label{ablation-complex}
\begin{adjustbox}{width=\linewidth,center}
  \begin{tabular}{@{}l|c|c|c|c|c@{}}
    \toprule
      \multicolumn{6}{c}{Hyperparameters}\\
    \hline \hline
         Approach & SR (\%) $\uparrow$ & AR (\%) $\downarrow$ & GFR (\%) $\downarrow$ & $d_{\text{total}}$ (m) $\downarrow$ & $v_{\text{total}}$ $\downarrow$ \\ \hline
    \hline
    Act\textit{Per}MoMa-Quality (0.7) & \textbf{91.4} & \textbf{0.6} & 8.0 & \textbf{4.57 $\pm$ 2.39} & \textbf{16.47$\pm$8.78}  \\ \hline
    Act\textit{Per}MoMa-Quality (0.9) & 88.8 & 5.0 & \textbf{6.2} & 4.84$\pm$2.73 & 17.11$\pm$9.54 \\ \hline\hline
    Act\textit{Per}MoMa-StableGrasp (1) & \textbf{92.6} & 2.2 & \textbf{5.2} & \textbf{4.57$\pm$2.51} & \textbf{16.20$\pm$8.65}  \\ \hline
    Act\textit{Per}MoMa-StableGrasp (5) & 91.2 & 2.8 & 6.0 & 4.92 $\pm$ 2.48 & 17.47$\pm$8.80  \\ \hline\hline
    Act\textit{Per}MoMa-IGweight (3.0) & 85.8 & 8.6 & 5.6 & 5.10$\pm$3.18 & 17.83$\pm$ 10.38  \\ \hline
    Act\textit{Per}MoMa-IGweight (0.2) & \textbf{92.6} & \textbf{1.8} & 5.6 & \textbf{4.31$\pm$2.27} & \textbf{15.60$\pm$8.39} \\ \hline \hline
    Act\textit{Per}MoMa-momentum (0) & 83.2 & 13.0 & \textbf{3.8} & 5.45$\pm$3.40 & 18.77$\pm$11.06\\ \hline 
    Act\textit{Per}MoMa-momentum (700) & \textbf{90.6} & \textbf{3.8} & 5.6 & \textbf{4.56$\pm$2.53} & \textbf{16.28 $\pm$ 9.11}  \\ \hline  
    \midrule
   \multicolumn{6}{c}{Ablation} \\ \hline \hline
    Act\textit{Per}MoMa$^\dagger$  & 92.6 & 1.80 & 5.6 & 4.31$\pm$2.27 & 15.60$\pm$8.39  \\ \hline
    Act\textit{Per}MoMa-IG-only & \textbf{96.8} & \textbf{0.8} & \textbf{2.4} & 3.52$\pm$1.61 & 12.87$\pm$6.31  \\ \hline
    Act\textit{Per}MoMa-no-weights & 48.4 & 42.0 & 10.6 & 9.05$\pm$3.79 & 13.28$\pm$10.44  \\ \hline
\midrule
    \multicolumn{6}{c}{Hard Grasps}\\ \hline \hline
    Act\textit{Per}MoMa$^\dagger$  & \textbf{61.8} & \textbf{29.6} & 8.6 & 7.19$\pm$4.17 & \textbf{24.81$\pm$13.24}  \\ \hline
    Act\textit{Per}MoMa-IG-only & 56.2 & 36.2 & \textbf{7.6} & \textbf{5.77$\pm$4.11} & 25.52$\pm$13.82 \\ 
    \bottomrule
  \end{tabular}
  \end{adjustbox}
  \end{center}
\vspace{-0.15cm}
   \scriptsize{$^\dagger$ Quality=0.8, StableGrasp=1, IGweight=0.2, momentum=800}
   \vspace{-0.48cm}
\end{table}

Tables~\ref{ablation-simple} and~\ref{ablation-complex} present the results of our study for simple and complex scenes, respectively. Focusing first on Table~\ref{ablation-simple}, we notice that in general for simple scenes the SR is high. That is because the robot has the freedom to explore the scene and find good grasps. From the hyperparameters, it is evident that even with a low grasp quality threshold we can achieve over 90\% SR with low GFR, but the rest of the metrics are comparable. We see that the temporal window during which a detected grasp is stable does not really change the performance. However, we notice that using too large time windows (over 10 frames as in~\cite{Breyer22}) leads to poor mobile grasping results. The use of a small weight on the IG once we detect grasps seems to play a critical role in performance, as it shows that once we find a grasp, we can consider exploiting this information while maintaining some exploration. Finally, in simple scenes, the momentum may sometimes even lead to a drop in performance, as it may end in more failed attempts. Continuing with the ablations, we see that the method that does not penalize the utilities with the path-related scaling performs very poorly. We note that VGN, trained on table-top environments, mainly favors top-down grasps that are easier to execute. 
Conversely, when we restrict VGN to generate only side ``hard'' grasps (at 45~º), we see a significant drop in performance in finding grasps~(higher AR). 

Table~\ref{ablation-complex}, with complex scenes, shows a similar trend but with a noticeable drop in the overall performance, which is evident also by the generally higher grasp failure rates. Here we see the benefits of our momentum term as high momentum in complex scenes leads to better performance and reduced path lengths. Our ablation places the method without the grasp-related utility higher than the full Act\textit{Per}MoMa approach, but in the ``hard'' grasp scenario, we notice a significant~$\sim$6\% performance improvement when accounting for the grasps utility while planning. Overall, we propose our full objective as it has a higher chance of giving us reliable base poses for executing mobile grasps, especially when considering more realistic real-world scenarios where many hard grasps also exist, e.g., due to a specific grasp-affordance that needs to be respected.

\subsection{Comparison with baselines}
We first consider baseline methods that do not use active perception, in the sense of using an information gain objective, as they are intuitive and can show in which cases active perception is necessary. We then consider a state-of-the-art method in the active grasping literature. Namely, we compare against (i) a naive approach (Naive) in which we navigate the robot towards the approximate target location and activate grasp execution if, within a 0.85~m distance from the object, a high-quality grasp has been detected; (ii) a random approach (Random) in which, at each time step, we randomly select a feasible base goal (i.e. no collision found with the currently perceived scene, also considering the reachability of the object from the base goal) around the approximate target object location. If a grasp has been detected, we execute the grasp. If the grasp is not successful, we resample a feasible goal with a smaller radius (0.75~m); and (iii) the method by Breyer et al.~\cite{Breyer22} adapted for mobile manipulation, in which we compute the IG per view (and not accumulated over paths) sampled on a hemisphere of radius 1m around the approximate object location. In this case, we always move to the viewpoint with the highest IG. If no grasp is found, we resample views with a smaller radius, and if we are within reach of the object and a grasp has been found, we execute it. In case of grasp failure, we move to the NBV. 

Table~\ref{baselines} presents the comparative results both for simple and complex scenes. Interestingly, both the Naive and Random approaches outperform the active grasping approach of Breyer et al.~\cite{Breyer22}. Thus, heuristic approaches like the Naive approach can work in simple scenes, alleviating the need for path planning. The landscape changes when looking at the results for the complex scenes. Act\textit{Per}MoMa still has the highest performance, while the naive approach now performs worse than Breyer et al.~\cite{Breyer22}. The random method performs well by moving close to the target and continuously sampling different base goals close to it. We posit that, as with Act\textit{Per}MoMa-IG-only, already sampling base goals that are collision-free according to the currently perceived scene and within the reachability radius of the robot is a strong inductive bias for planning successful robot placements for robotic grasping. Nevertheless, as with the previous experiments in sec.~\ref{sec:ablation}, the hard-grasp case shows the significant benefit of Act\textit{Per}MoMa w.r.t baselines. Notably, we highlight the large benefit in finding grasps (at least~$\sim$20\% lower abort rate) compared to baselines. 

\begin{table}[t!]
\renewcommand{\arraystretch}{0.9}
\centering
\captionsetup{justification=centering}
  \caption{Comparison with baselines}  \label{baselines}
\begin{adjustbox}{width=\linewidth,center}
  \begin{tabular}{@{}l|c|c|c|c|c@{}}
    \toprule
    \textbf{Approach} & \textbf{SR (\%)} $\uparrow$ & \textbf{AR (\%)} $\downarrow$ & \textbf{GFR (\%)} $\downarrow$ & ${d}_{\text{\textbf{total}}}$ \textbf{(m)} $\downarrow$ & $v_{\text{\textbf{total}}}$ $\downarrow$ \\ \hline \hline
    \multicolumn{6}{c}{Simple scenes} \\ \hline \hline
    Naive & 95.2 & \textbf{1.2} & 3.6 & \textbf{1.36$\pm$0.28} & \textbf{5.93$\pm$3.77}  \\ \hline
    Random & 93.2 & 5.0 & 1.8 & 4.38$\pm$1.98 & 15.24$\pm$6.47  \\ \hline
    Breyer et al.~\cite{Breyer22} & 92.0 & 8.0 & \textbf{0.0} & 3.71$\pm$1.78 & 12.56$\pm$6.12  \\ \hline
    Act\textit{Per}MoMa (ours) & \textbf{95.4} & 1.4 & 3.2 & 3.59$\pm$1.69 & 12.67 $\pm$ 5.39 \\ \hline
    \midrule
    \multicolumn{6}{c}{Complex Scenes}\\
    \hline \hline
    Naive & 86.8 & 2.6 & 10.6 & 1.55$\pm$0.45 & \textbf{8.72$\pm$8.89}  \\ \hline
    Random & 90.4 & 5.8 & \textbf{3.2} & 3.97$\pm$1.56 & 13.98$\pm$5.39  \\ \hline
    Breyer et al.~\cite{Breyer22} & 90.0 & 6.0 & 4.0 &\textbf{ 3.32$\pm$1.47} & 12.32$\pm$6.13  \\ \hline
    Act\textit{Per}MoMa (ours) & \textbf{92.6} & \textbf{2.2} & 5.2 & 4.57$\pm$2.51 & 16.20$\pm$8.65  \\ \hline
    \midrule
    \multicolumn{6}{c}{Complex Scenes (Hard Grasps)}\\
    \hline \hline
    Naive & 43.8 & 49.2 & 7.0 & 6.06$\pm$3.81 & 27.25$\pm$13.95  \\ \hline
    Random & 24.6 & 68.4 & 7.0 & \textbf{3.82$\pm$2.23} & \textbf{16.29$\pm$11.07}  \\ \hline
    Breyer et al.~\cite{Breyer22} & 47.2 & 48.8 & \textbf{4.0} & 5.11$\pm$3.83 & 24.03$\pm$14.88  \\ \hline
    Act\textit{Per}MoMa (ours) &  \textbf{61.8}  & \textbf{29.6} & 8.6 & 7.19$\pm$4.17 & 24.81$\pm$13.24  \\ \hline
    \bottomrule
  \end{tabular}
  \end{adjustbox}
  \vspace{-0.55cm}
\end{table}

\subsection{Real robot demonstration}


To evaluate the applicability of Act\textit{Per}MoMa we conduct real-world experiments in a cluttered room, as shown in Fig.~\ref{fig:real_tiago}. We ran 10 experiments, re-arranging the setup every time and selecting a different target object. The target object is placed on the table while the rest of the scene objects act as obstructions. The grasp quality threshold is increased to 0.9, higher than in simulation, to account for the sim-to-real gap for the grasping network. We achieved real-world performance with an 80~\% success rate, 20~\% aborts, and no grasp failures, with an average distance traveled of $3.09 \pm 1.06$~m, and $13.50\pm13.86$ number of views. Additional details and demonstrations are provided in the supplementary video and the project website: \href{https://sites.google.com/view/actpermoma/home}{https://sites.google.com/view/actpermoma}.

\subsection{Limitations} An issue of methods that use reactive planning (such as ours) is that, depending on the resolution of the sampled base goals and the sampling frequency, the robot can get stuck in deadlocks trying to switch between base goals leading to oscillating motions. Although we introduce a penalty for this behavior, namely the momentum, some amount of deadlocks due to changes in direction can still exist. Another limitation is the limited volumetric information in the target area in very occluded scenes, making the IG computation difficult. This can prominently be observed in Breyer et al.~\cite{Breyer22}, as very occluded objects that get partially discovered from different views often prompt a `zigzag' path. We improve this behavior by not just considering the best NBV to decide where to go but instead using the whole spatial distribution provided by our sampled base goals. A possible mitigation for this in future work is to train a reinforcement learning agent on these POMDP problems, leveraging a combination of our active and some random exploration.

While the real-world demonstration shows promise toward autonomous perceptive MoMa, some practical challenges remain. Most prominently, the camera frame rate and RGBD integration can significantly limit the algorithm's control frequency (Act\textit{Per}MoMa frequency is about 10~Hz, but can drop to 5~Hz in the real world). Secondly, continuous stable grasp detection while moving can be challenging, especially with imperfect localization.

%% file: 4_conclusion.tex
\section{Conclusion}
In this paper, we delved into the intricate challenges of active perception for mobile manipulation in unknown and cluttered environments. We introduced a novel formulation that generates robot paths toward objects of interest without any prior scene knowledge, drawing upon principles of active perception and mobile manipulation coordination. Our approach seamlessly combines exploration — for maximizing visual information gain---and exploitation of task-specific parameters such as grasp executability. Using a receding horizon control strategy, we ensure the robot's motion can adapt dynamically to new data. Our experiments with the dual-arm TIAGo++ mobile manipulation robot have further validated the feasibility and efficiency of our proposed method in cluttered environments. While our results have shown how active perception can provide performance gains for efficient mobile grasping, the vast potential of active perceptive mobile manipulation remains uncharted territory. Looking ahead, we aim to explore deep learning techniques to predict scene information gain to use in robot learning for mobile manipulation to tackle even more challenging tasks.